\documentclass[runningheads]{llncs}
\usepackage{amsmath}
\usepackage{graphicx}

\usepackage{xcolor}

\begin{document}
\sloppy

\title{Using Deep Reinforcement Learning Methods for Autonomous Vessels in 2D Environments}

\titlerunning{Vessel Navigator with Planner and Local View (VNPLV)}

\author{Mohammad Etemad\inst{1}
\and Nader Zare\inst{1}\and Mahtab Sarvmaili\inst{1} \and Am\'ilcar Soares \inst{1} \and Bruno Brandoli Machado\inst{1} \and Stan Matwin\inst{1}\inst{2}
}

\institute{
Institute for Big Data Analytics, Dalhousie University, Halifax
\and
Institute for Computer Science, Polish Academy of Sciences, Warsaw}

\authorrunning{M. Etemad et al.}

\maketitle             
\begin{abstract}
Unmanned Surface Vehicles technology (USVs) is an exciting topic that essentially deploys an algorithm to safely and efficiently performs a mission. 
Although reinforcement learning is a well-known approach to modeling such a task, instability and divergence may occur when combining off-policy and function approximation. 
In this work, we used deep reinforcement learning combining Q-learning with a neural representation to avoid instability. 
Our methodology uses deep q-learning and combines it with a rolling wave planning approach on agile methodology. 
Our method contains two critical parts in order to perform missions in an unknown environment. 
The first is a path planner that is responsible for generating a potential effective path to a destination without considering the details of the root. 
The latter is a decision-making module that is responsible for short-term decisions on avoiding obstacles during the near future steps of USV exploitation within the context of the value function. 
Simulations were performed using two algorithms: a basic vanilla vessel navigator (VVN) as a baseline and an improved one for the vessel navigator with a planner and local view (VNPLV). 
Experimental results show that the proposed method enhanced the performance of VVN by 55.31\% on average for long-distance missions. 
Our model successfully demonstrated obstacle avoidance by means of deep reinforcement learning using planning adaptive paths in unknown environments.

\keywords{Deep reinforcement learning \and path planning \and obstacle avoidance \and maritime autonomous surface vessels}

\end{abstract}

\section{Introduction}
\label{sec:introduction}

Ship collision avoidance and path planning is a fundamental research topic for autonomous navigation. 
Several methods have been proposed in the literature to this end. 
However, deep reinforcement learning strategies have empowered models for automatic maneuverability of vessels \cite{CHENG2018DLObstacleAv}. 
Autonomous vehicles have been developed and improved in different areas, ranging from unmanned aerial (e.g., planes and drones) \cite{kanistras2015survey}, to underwater (e.g., ships, remotely operated underwater vehicles (ROVs), submarine gliders, and unmanned suface vessels (USVs)) \cite{WYNN2014UUVsurvey}.

Path planning has the objective of generating a path between an initial location and the desired destination with an optimal or near-optimal performance under specific constraints. 
Avoiding obstacles in real or simulated environments is an essential task for safely driving USVs towards a target without human intervention \cite{Elkins2010AMaritimeNav}. 
For example, in marine application scenarios, it is of extreme importance to avoid obstacles such as rocks, floaters, debris, and other ships \cite{CHENG2018DLObstacleAv}.

In this paper, we develop a new method for path planning and obstacle avoidance in marine environments by using deep reinforcement learning (DRL) and local view strategy, namely Vessel Navigator with Planner and Local View (VNPLV). 
Unlike previous method as proposed in \cite{wang2018path}, which has a single environment with a fixed origin and destination points, we developed a methodology that can surpass traditional global approaches in unknown environments without any limitation of various origin and destination points. 
Basically, we improve the performance of the model by feeding it with CNN features and reducing the number of states using the Ramer–Douglas–Peucker algorithm. 
Results show that we can further enhance our model by using the idea of rolling wave planning and that our method benefits from combining the path planner for longterm planning and local view for short term decisions \cite{larman2004agile}.

We summarize the contributions of our work as follows. Inspired by agile methodology, we implemented the idea of rolling wave planning by using a deep reinforcement learning model for short-term decision making and a planner for longterm general planning.  
    ii) We created 2D marine environments by extracting information of geographical layers.
    iii) We developed a deep reinforcement learning method for agents that simulates USV movements. Agents trained by our method are able to autonomously plan their path and avoid obstacles in a simulated 2D marine environment. 
    iv) We performed extensive experiments by means of simulations comparing our method and a baseline and propose a metric named Rate of Arrival to Destination (RATD) to evaluate the performance of our method.

The rest of this paper is organized as follows.
Section \ref{sec:relatedworks} presents some related works in the area of path planning, reinforcement learning, and deep reinforcement learning.
In Section \ref{sec:methods}, we provide definitions used across our work and detail our four proposed methods. 
Section \ref{sec:experiments} demonestrates the performance of our method in a simulated marine environment. 
Finally, Section \ref{sec:conclusions} concludes the work and also discusses future works.

\section{Related works}
\label{sec:relatedworks}


In recent years USVs have attracted a great deal of attention from several maritime companies and research groups all over the world. 
There exist several approaches developed and applied for the USVs, which are mainly divided into path planning, obstacle avoidance, and intelligent optimization methods. 
We divide this section into three subsections to which our work is related. 

\textbf{\emph{Traditional path planning approaches.}} Several research methods on collision avoidance have been developed for path planning based on A* and Artificial Potential Field (APF). The algorithm A* is a global heuristic search strategy that takes into account both the start position and the destination. However, the algorithm is considered in the literature to be an inefficient search in a large grid map. The works of \cite{cheng2014} and \cite{wang2014} proposed hierarchical path planning strategies to improve the efficiency of A*. In contrast to A* algorithm, APF employs repulsive fields to model the environment with higher efficiency \cite{lyu2018fast}. Such a strategy can create a smoother path when it compares with A*.

\textbf{\emph{Reinforcement learning.}} Some articles have employed reinforcement learning (RL) for the collision avoidance and path planning task to improve the autonomy of the obstacle avoidance system. 
Reinforcement learning is a classical machine learning method, first proposed by Sutton in 1984, and widely explored in the '90s. It has been widely used in the artificial intelligence field. 
Although the main algorithm used for path planning is Q-Learning \cite{shen2019}, many methods in the literature are hybrid of RL with other methods \cite{Magalhaes2018}. 
In general, the two main drawbacks of RL approaches are the high cost of the learning process that depends on the environment and the user condition, and the degrees of freedom (DOF). Although the reinforcement learning algorithms have shown successful performance in variety of domains, their applicability has been  restricted to fully observable low-dimensional state spaces domains. 

\textbf{\emph{Deep reinforcement learning.}} DRL is a novel topic which has been emerged to address the challenges of using RL in complex, high dimensional problems. In addition to the outstanding performance of Deep RL models in other domains, they have attracted a lot of attention in ship collision avoidance topic. The work of \cite{CHENG2018DLObstacleAv} proposed a deep reinforcement learning obstacle avoidance approach with the deep Q-network architecture for unmanned marine vessels in unknown environment. The authors presented a learning policy for obstacle avoidance at a safe distance in unknown environments with 3-DOF. They used the replay buffer and self-play trials to learn the control behaviors. Recently, \cite{shen2019} presented a prototype of multiple autonomous vessels controlled by deep q-learning. Their reward function and the training process were designed with respect to the maneuverability of the vessel, including speed, and acceleration. The incorporated navigation rules employed the conversion to navigational limitation by polygons or lines.

It is also important to point out that there exist two kinds of analysis regarding the enviroment exploration: (i) the global view, and (ii) the local view. Environment exploration is linked to the performance of the approaches, but there is no strict definition related to the size of the local view. The most noteworthy approach for local view strategy is based on line-of-sight (LOS), presented by \cite{tran2013}. Moreover, utilizing a local view became a best practice in some strategic game solvers such as \cite{vinyals2019grandmaster}.

In this paper we present a path planning method using deep q-learning with unknown position of dynamically generated obstacles in the environment. 
Our DRL-based method is focused on a local view strategy which reduces the number of states. Our policy is obtained using four different iterations of evolving our proposed work evaluated in simulation experiments detailed in Section \ref{sub:pplaning}.

\section{A New DRL Method for Unmmaned Surface Vessels}
\label{sec:methods}

In this section we go through the details of the proposed method. 
First, we define the main concepts used to model our agent-environment approach (Section \ref{subsec:defs}). 
After that, we describe our proposed method for path planning and obstacle avoidance in the maritime domain (Section \ref{sub:pplaning}).

\subsection{Definitions}
\label{subsec:defs}

In this work, an \textit{\textbf{agent}} is a vessel voyaging from an origin in the environment with the objective of safely arriving at a desired destination in the environment. 
The \textit{\textbf{environment}} is a bounding box area that contains a body of water that an agent can voyage through and variety of lands which the agent cannot travel through.
The environment has access to a layer of obstacles, such as vessels moving in the environment.
Our agent can take some actions from a set of directions to move from its current location to its next location in a direction for 0.001 degrees, which is about 100 meters of distance traveled in the real world.

An \emph{\textbf{origin point}} is the position of an agent at the first moment of the training or testing phases of our methods. 
This means that our agent is positioned at the location of the origin point at the start of each experiment or training episode. We can define an origin point, $OP=(x_s,y_s)$, as a tuple where $x_s$ is the latitude of the agent at the start of the experiment or training episode, and $y_s$ is the longitude of the agent.

The \emph{\textbf{destination point}} is the final geographical position that an agent should arrive at.  
We define $DP=(x_d,y_d)$ as a tuple where $x_d$ is the latitude of the desired location that the agent aims to arrive at, and $y_d$ is the longitude of that location.

We use eight discrete \emph{\textbf{actions}} $A = \{N, S, E, W, NE, NW, SE, SW\}$, representing the directions North, South, East, West, Northeast, Northwest, Southeast, and Southwest, respectively.

In this work, we use five discrete \emph{\textbf{outcomes}} for an action is taken by an agent $O = \{$\emph{hit an obstacle, hit land, arrive at target, vanish target, normal movement}$\}$.
To \emph{\textbf{hit an obstacle}} means that the agent hits one of the vessels moving in the environment. 
To \emph{\textbf{hit land}} means the agent took a direction that moved it to a geographic area that has land, which is not suitable for the vessel. 
To \emph{\textbf{arrive at target}} means that our agent successfully reached its destination point. 
To \emph{\textbf{vanish target}} means that the agent is farther from the target than the distance threshold. 
Finally, a \emph{\textbf{normal movement}} is an output where the agent has not finished its mission, but there is no reason to stop the voyage.

Each action and its outcome for our agent in the environment is considered as one \emph{\textbf{step}} $s_i = (a_i, o_i)$, where $s_i \in S$, $a_i \in A$, $o_i \in O$, and $S$ is the set of all possible steps. 
Therefore, each step moves our agent from a current state to a future state.

An \emph{\textbf{episode}} is a set of consecutive steps with a fixed origin and destination point. 
In an episode, the agent voyages from the origin point with the objective of arriving at its destination point. 

An episode $e_j = (OP_j, DP_j, <s_1,s_2,...,s_n>)$ - where $OP_j$ is an origin point, $DP_j$ is its destination point,  $<s_1,s_2,...,s_n>$ - is a sequence of steps, and $e_j \in E$, where $E$ is the set of all possible episodes. 
The outcome of an episode is the outcome of the last step of that episode, which is $s_n$. 
When the outcome of an episode is to \emph{hit an obstacle}, \emph{go to land}, \emph{vanish target}, this is considered to be a \emph{\textbf{failed episode}}. 
If the outcome of an episode is \emph{arrive at target}, this is considered to be a \emph{\textbf{successful episode}}. 
In this work, we define a \emph{\textbf{maximum number of steps}} for each episode because we want to encourage our agent to arrive at its destination as fast as possible and to avoid repetitive actions.
If the number of steps in each episode exceeds this number, we call that episode a failed episode as well.

A \textbf{\emph{plan}}, which is defined as $p_k = (OP_k, DP_k, <e_0, e_1,..., e_m>)$ where, $OP_k$ is an origin point, $DP_k$ is its destination point, and $<e_1,e_2,...,e_m>$ is a set of episodes. 
The destinations in each episode ($e_0, e_1, ..., e_m$) of a plan $p_k$ are called \textbf{\emph{intermediary goals}}, except for the last episode, which is the final destination reached by a plan. 
If the outcome of a plan is not reached the destination, we call that plan a \emph{\textbf{failed plan}}. 
The \emph{arrive at target} outcome means the agent has (i) arrived at target and (ii) the agent is in its destination. 
If the outcome of a plan is to reach the destination, we call that plan a \emph{\textbf{successful plan}}. 

\begin{figure}[b]
    \centering
    \includegraphics[width=11cm]{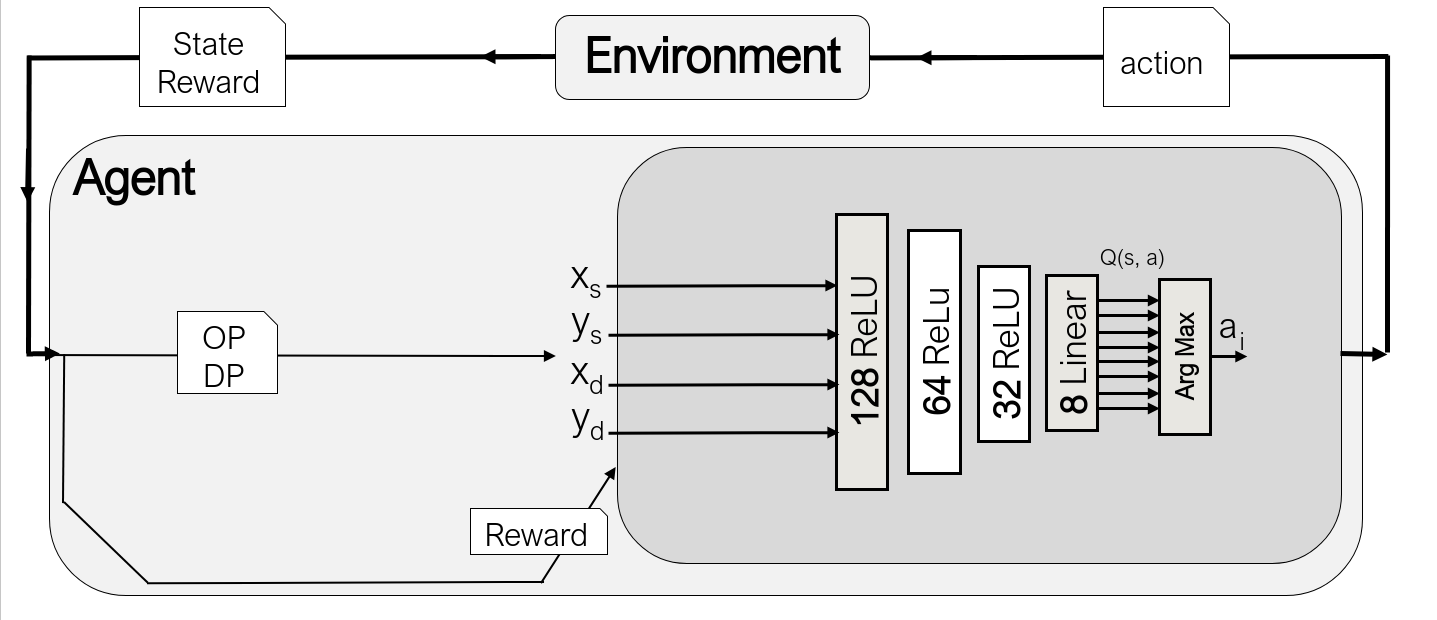}
    \caption{Vanilla Vessel Navigator (VVN)}
    \label{fig:VVN}
\end{figure}
We limit the knowledge our agent has about the environment so that the agent is only able to observe within the boundary around itself. 
This approach of creating a local view has been used in \cite{vinyals2019grandmaster}; however, the local view of our work is not a limitation for an agent. 
In this way, we force the agent to learn general rules of movement, without memorizing the whole environment and the best paths. 
Furthermore, some details in the environment, such as dynamic obstacles that are far from an agent, can move to other locations by the time our agent arrives there. 

Having access to full information about the environment would encourages our agent to memorize the environment.
When an intermediary goal is outside of our agent's local view, a subset of our environment, we make an abstract line from the agent position to its intermediary goal and find the shadow of the intermediary goal inside the local view with some margin of freedom.

\subsection{Baseline and our proposed method}
\label{sub:pplaning}

\begin{figure}[tb]
    \centering 
    \includegraphics[width=11cm]{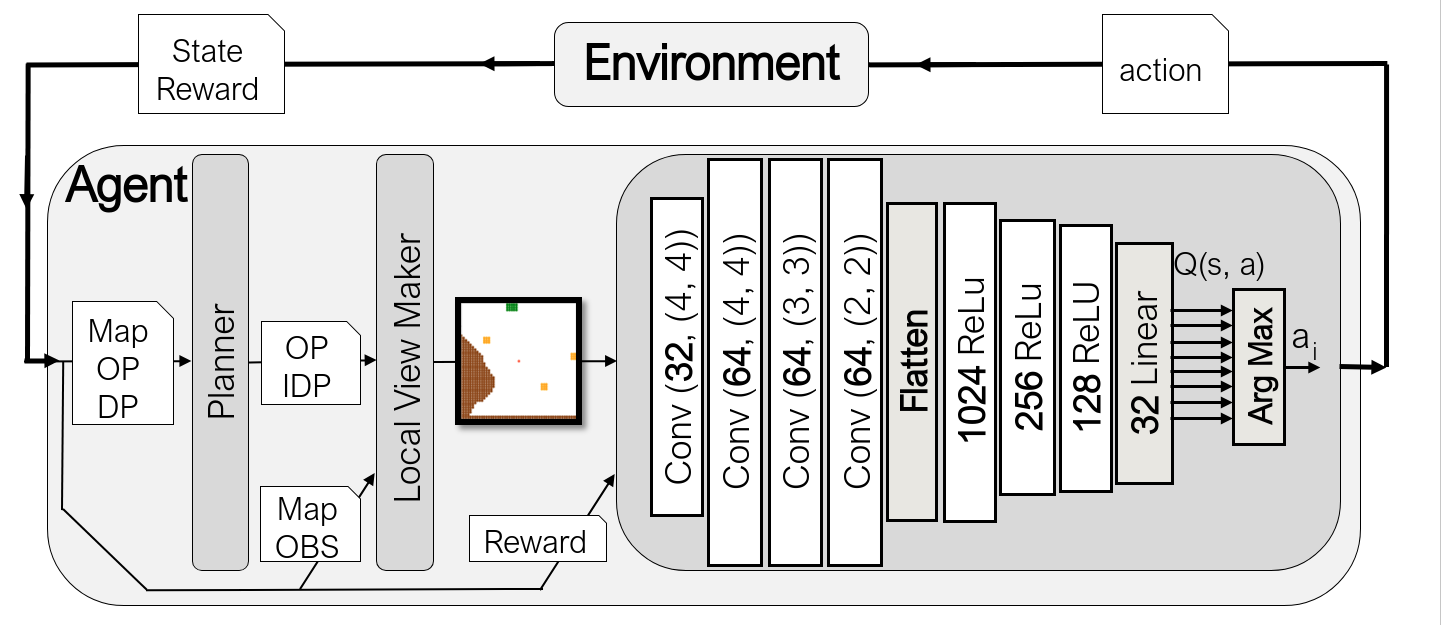}
    \caption{Vessel Navigator with planner and localview (VNPLV)} \label{fig:VNPLV}
\end{figure}

In Figures \ref{fig:VVN} and \ref{fig:VNPLV}, we summarize the baseline method and our proposed one, respectively.  
In these figures, $x_s$ and $y_s$ are the origin point coordinates, $x_d$, $y_d$ are destination point coordinates, $a_i$ is the action suggested by the method, $Q(s, a)$ is the action value function that predicts the reward given to an agent if it selects an action $a$ in state $s$, $ReLU$ is a rectified linear unit, $Linear$ is a linear activation function, $Flatten$ is a flattening process that involves transforming the entire pooled feature map matrix into a single column, and  $Conv$ is a convolutional neural network. 
We detail how these methods work below.

\textbf{Vanilla Vessel Navigator (VVN).} 

This method was implemented based on the work of \cite{wang2018path}, but we addressed the two limitations of their algorithm for a fair comparison with our method. 
The first limitation is that the model's destination is a static point, and as a result, the model is able to learn only routes to a single destination. 
The second limitation is that using the Q\-Learning approach with a large and dynamic environment makes it impossible for Q\-Learning to train.

In our implementation of the VVN method, the agent receives its origin and destination point with freedom inside the environment.

Figure \ref{fig:VVN} shows the architecture of this method where the origin point ($x_s$ and $y_s$), and destination point ($x_d$ and $y_d$) are the only inputs that feed our model. 
The output of this architecture, which is the $argmax$ of the outputs of our neural network, determines the action of our agent.
Since the objective of this paper is not to search for the best model for this architecture, we select one model with reasonably good results on training using a trial-and-error approach.

In the original model presented by \cite{wang2018path}, by changing the 
destination point, the agent would need to be trained again. 
Therefore agent trained by the original method is not able to use its knowledge from previous training.
Our modification removed this limitation so that the origin and destination points can be selected dynamically, and such retraining limitations are not necessary.
Figure 1 shows the reactions between the agent and the environment. The agent receives two origin and destination points, OP and DP, from the environment and the reward that the agent gained during its last action. The neural network model estimates the next action of the agent by updating its parameters using the reward of the previous action.

\textbf{Vessel Navigator with Planner and Local View (VNPLV).} 
 
In the first step of our proposed method (Figure \ref{fig:VNPLV}), we introduce a path planner that makes a full plan from the start point to the destination without considering any dynamic obstacle in its way using the Floyd Algorithm \cite{floyd1962algorithm}.  
The Floyd algorithm has the objective of finding the shortest paths in a weighted graph. This algorithm is computationally expensive but we only run it once for an environment and store the calculations results.
This can be seen as high-level planning without considering the details of a plan. 
In the second step, we reduce this high-level plan by removing similar intermediary goals applying the Ramer–Douglas–Peucker algorithm \cite{hershberger1994n}, in our work $\epsilon=0.001$ geomtry degree.
The Ramer–Douglas–Peucker algorithm has the objective of simplifying a curve composed of a line of segments to a similar curve with fewer points.
This reduction gives our agent more flexibility in making local decisions. 
In the third step, our agent decides on the details of the plan in the near future to arrive at the shadow of its intermediary goal residing in the local view. The shadow is a point in the direction to intermediary goal inside a definde margin of the local view.in This work we use a margin of 3 pixcels. 
Our agent observes its local view and finds an abstract destination point, which is the shadow of the nearest intermediate destination provided by the path planner. 
This first destination point is the destination point of the first episode in this plan, which resides in the local view.  
The agent starts moving towards its abstract destination point, which is residing inside the local view.
After achieving the intermediary goals (i.e., the destination point of that episode), the environment updates the destination point of the next episode in its plan. 
The idea of using CNN is not a novel idea and has been introduced in reinforcement learning by the work of \cite{mnih2015human}. Moreover, the idea of using a local view also is introduced in \cite{vinyals2019grandmaster}. To the best of our knowledge, we are the first to use a combination of path planning and local view for implementing the rolling wave planning approach for this problem. 
In Figure 2, our agent benefits from two planning modules. First, the path planner provides a longterm plan without considering any dynamic obstacles on the way to arriving at the destination. 
Second, a DRL decision-making approach provides details for short-term decisions actions in detail. This part is responsible for avoiding obstacles and moving the agent safely in the near future.

\section{Experiments}
\label{sec:experiments}

In this section we describe the dataset and evaluation metrics (Sec. \ref{sub:dataset}), the setup for our algorithm (Sec. \ref{sub:setup}) and the experiments performed (Sec. \ref{sub:analysis}). 
\subsection{Dataset creation and evaluation metrics}
\label{sub:dataset}

We selected a region in the area of Halifax, Nova Scotia (Canada) with a bounding box (\emph{longitude}, \emph{latitude}) starting from ($-63.69, 44.58$) and ending at ($-63.49, 44.73$).
We created a water and a land layer using public data on earth elevation using the  NOAA\footnote{https://coast.noaa.gov/dataviewer/} dataset. 
 
Then we drew a buffer of 50 meters around the land and used it as the environment. 
We also randomly added some moving objects to play the role of obstacles in this simulated 2D environment. 
In this work, we dynamically generated the obstacles during an episode. 
However, the design we have in mind for the future is based on the assumption that these obstacles can be dynamic and can move based on traffic patterns, such as can be extracted from Automatic Identification System (AIS) messages.

We also defined a metric called RATD to measure the performance of a method as follows. 
When we test a method, we randomly generate a set of origin and destination points $N$. 
Such information is provided to the method being tested as an episode or a plan. 

Then we observe if the method can successfully place the agent at its destination (i.e., a successful episode or a successful plan) or not (i.e., a failed episode or a failed plan). 
We count how many times a model successfully conducts the agent to its destination and call it $P$. 
The RATD is calculated as $RATD=\frac{{|P|}}{{|N|}}*100$.

\subsection{Training and testing setup}
\label{sub:setup}

In our experiments, we use the reward function $R(o_i \in O)$ detailed in Equation \ref{eg:1}, where $\Delta_d$ is the distance in geometry degree that an agent moved from the last state, $\psi$ is set at 1,000, $\Delta_{od}$ is the distance of our agent from its nearest obstacle in geometry degree, and $\phi$ is set at 20.
In Equation \ref{eg:1} we deducted $\kappa$ from the reward to encourage the agent to find the nearest path, and it was set at 0.01. 
The values of $\psi$, $\phi$, and $\kappa$ were manually tuned using a trial and error approach and the values reported are those that provided the optimal performance for all methods.

\begin{equation}
R(x)= \begin{cases}
    -5,              & \text{Vanish target, Obstacle collision, Land collision }\\
    +5,              & \text{Arrive target}\\
    \psi*\Delta_d - \phi*\Delta_{od} - \kappa,              & \text{Other}\\
\end{cases}
\label{eg:1}
\end{equation}

In this work, we use a \textit{target network} to adjust the action-values (Q) iteratively towards the target values as in \cite{mnih2015human}. 
The application of target network application is to reduce correlations with the target.
In the work of \cite{mnih2015human}, a reply buffer is proposed to eliminate relationships in the observation chain and to soften fluctuations in the data distribution, and we use the same idea here.

In our training phase, a \textit{training step} includes 1,000 episodes or plans randomly executed to update our neural network parameters. 
Each of these episodes or plans includes an undefined number of training steps. 
We store i) the agent's state, ii) the agent's next state iii) the reward, and iv) the action, in a reply buffer with a size of 100,000 in the same way as introduced in \cite{mnih2015human}.

We retrieve 3,000 items from this reply buffer after every 100 training steps. 
Every 200 training steps, the parameters of our model are copied to our target network.
These values were configured empirically by trial and error tests.

In the training phase of our models we use the idea of exploit and explore which means the action of our agent is based on two types of learning \cite{brafman2002r}. 
First, exploring the environment randomly and measure the gained rewards. 
Second, exploiting the learned knowleadge by using the parameters of the trained neural network.
In this work, we assigned the weight of 0.9 for the explore part and 0.1 weight for the exploit part.
The weight of explore increases during the training so that in traing step 25,000 of our training, the weight of exploit becomes 0.9 and the weight of explore becomes 0.1. 
After training step 25,000, these weights are not changed.

\subsection{Result analysis and discussion}
\label{sub:analysis}

An experiment was conducted to answer the two following  research questions. 
First, what is the rate of arriving at the destination if we randomly select some sets of origin and destination points for episodes? 
Second, how does this rate change by increasing the distance between the origin and destination points?
We increased the distance between the origin and destination points using the following distances in degrees [$0.01,0.02, 0.04,0.08,0.16,0.32$], which is roughly equal to [$1364.59 ,2729.11, 5457.90, 13004.68,21824.01, 43627.70$] in meters.

\begin{figure}[tb]
        \centering
        \includegraphics[width=1.\textwidth]{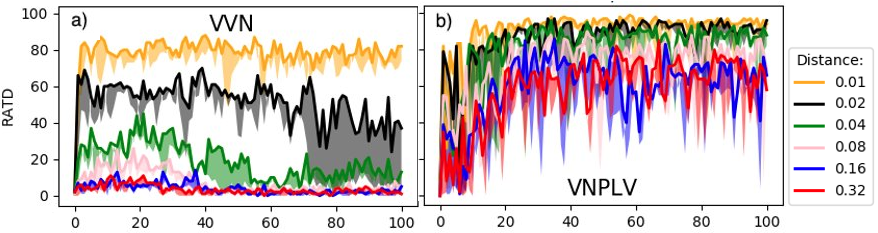}

        \caption[ The average and standard deviation of critical parameters ]
        {\small Results of training of each model for different sets of distances between the start point and destination.} 
        \label{fig:train_res}
\end{figure}

Figure \ref{fig:train_res} shows the results of our experiment for each training step. 

Therefore, each point in Figure \ref{fig:train_res} represents the results of 100 tests while the x-axis shows the progress on the number of training steps in a $10^2$ scale. 
Between each two points, we update the parameters of our models using 1,000 executions of episodes or plans so that they update the parameters of our network.
In this experiment, we dynamically randomly add some vessels on the body of water to play the role of obstacles.

The results for VVN (Figure \ref{fig:train_res}(a)) show that by increasing the distance between the origin point and the destination point, the percentage of successful trips decreases. In the experiment with a distance of 0.32 degree from origin to destination, the mean of RATD decreased from 79.44\% (using VNPLV) to 24.13\% (using VVN).
These results show a weakness of VVN, as it is only good for near distance situations and cannot perform well in long-distance path planning.
This is because (i) the search space of the agent increases when the distance is longer, and (ii) the probability of selecting actions from a loop of actions can increase so that the agent can just move back and forth without arriving at its destination.
In Figure \ref{fig:train_res}(a) the shortest distance, shown by yellow, achieved the highest RATD. As can be seen, by increasing the distance, the VVN performance declined so that the mean of RATD was 24.13\% for a 0.32 degree distance.

Figure \ref{fig:train_res}(b) shows the result for the VNPLV.  
 
The results shows that the performance of VNPLV is improved considerably, 55.31\% on mean of RATD for a 0.32 degree distance, in comparison to VVN. 
The proposed method is more stable and learns better to navigate our agent, even with fewer episodes for training. This enhancement is because VNPLV is equipped with two crucial modules to benefit from rolling wave planning. First, the longterm planner provides a potential optimal path without considering the details of the movement. Second, the local view makes the decisions related to navigating the agent in the near future to avoid dynamic obstacles.

\begin{figure}[t]
    \centering
    \includegraphics[width=.60\textwidth]{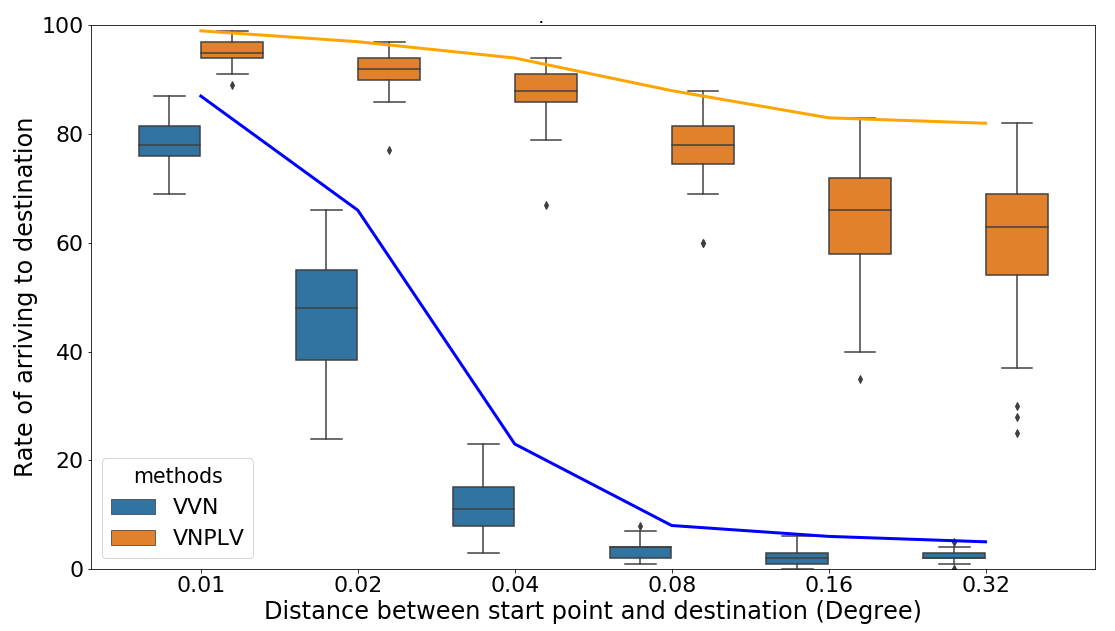}
    \caption{Comparing the two vessel navigation methods using RATD for six category of distances after training models for 100,000 training steps}
    \label{fig:compare}
\end{figure}

Figure \ref{fig:compare} presents a high level comparison of all methods developed in this work. 
Figure \ref{fig:compare} shows boxplots of the RATD for the distances of 0.01, 0.02, 0.04, 0.08, 0.16, and 0.32 degrees. 
As can be seen, VNPLV has the best performance values for all categories of distances.
We used a T-test to compare the results of VVN and VNLPV methods which supports that there is a statistically significant difference between VVN and VNPLV p\_value=0.000019.
The results show that the mean of the RATD for these two methods are statistically different, confirming that the performance of VNPLP is higher than the VVN for this experimental setup.

\section{Conclusions and Future Works}
\label{sec:conclusions}


In this work, we proposed a method to improve the performance of the RL Vessel Navigator model using the concept of rolling wave planning in agile methodology.

Our proposed method takes advantage of two planning approaches: 1) long-term planning using a path planner with the assumption of no obstacles to generate a potential efficient path, and 2) a short-term decision-maker that is the output of our reinforcement learning model to avoid dynamically generated obstacles and to navigate the agent in the near future.

Our experiments show that the use of a local view improves the performance of our basic model. 
However, its performance reduces when the distance between the origin point and destination increases. 
We address this weakness using a path planner to provide a potentially efficient path for the whole trip without the detail of movements. This is followed by a short-term decision-maker to navigate agents safely.

Although we applied the idea of adaptive planning from the agile methodology for autonomous navigation, it can be applied to any other domain to increase the agent's performance.
We intend to expand our work in various ways. 
We want to connect our dynamically generated obstacles to AIS messages received from vessels and test our agent in an unknown environment with dynamic obstacles. 
In this way, we would like to change the core neural network with sequence model networks such as a combination of RNN and CNN to have a memory of the past agent actions.
We also would like to add similar agents to the environment and define a global task to perform it simultaneously in the environment. 

\subsubsection*{Acknowledgments}
The authors would like to thank NSERC (Natural Sciences and Engineering Research Council of Canada) for financial support, and Jennifer Strang, GIS Analyst at the Dalhousie University Libraries, for her help in preparing the geographic data for analysis.
Computations were performed on the DeepSense computing platform. DeepSense is funded by ACOA, the Province of Nova Scotia, The Centre for Ocean Ventures \& Entrepreneurship (COVE), Ocean Frontier Institute (OFI), IBM and Dalhousie University.

\bibliographystyle{splncs04}
\bibliography{refs}

\begin{thebibliography}{10}
\providecommand{\url}[1]{\texttt{#1}}
\providecommand{\urlprefix}{URL }
\providecommand{\doi}[1]{https://doi.org/#1}

\bibitem{WYNN2014UUVsurvey}
et~al., R.B.W.: Autonomous underwater vehicles (auvs): Their past, present and
  future contributions to the advancement of marine geoscience. Marine Geology
  \textbf{352},  451 -- 468 (2014)

\bibitem{brafman2002r}
Brafman, R.I., Tennenholtz, M.: R-max-a general polynomial time algorithm for
  near-optimal reinforcement learning. Journal of Machine Learning Research
  \textbf{3}(Oct),  213--231 (2002)

\bibitem{cheng2014}
{Cheng}, L., {Liu}, C., {Yan}, B.: Improved hierarchical a-star algorithm for
  optimal parking path planning of the large parking lot. In: 2014 IEEE
  International Conference on Information and Automation (ICIA). pp. 695--698
  (July 2014). \doi{10.1109/ICInfA.2014.6932742}

\bibitem{CHENG2018DLObstacleAv}
Cheng, Y., Zhang, W.: Concise deep reinforcement learning obstacle avoidance
  for underactuated unmanned marine vessels. Neurocomputing  \textbf{272},  63
  -- 73 (2018)

\bibitem{Elkins2010AMaritimeNav}
Elkins, L.e.a.: The autonomous maritime navigation (amn) project: Field tests,
  autonomous and cooperative behaviors, data fusion, sensors, and vehicles.
  Journal of Field Robotics  \textbf{27}(6),  790--818 (2010).
  \doi{10.1002/rob.20367}

\bibitem{floyd1962algorithm}
Floyd, R.W.: Algorithm 97: shortest path. Comm. of the ACM  \textbf{5}(6), ~345
  (1962)

\bibitem{hershberger1994n}
Hershberger, J., Snoeyink, J.: An o (n log n) implementation of the
  douglas-peucker algorithm for line simplification. In: Proceedings of the
  tenth annual symposium on Computational geometry. pp. 383--384 (1994)

\bibitem{kanistras2015survey}
Kanistras, K., Martins, G., Rutherford, M.J., Valavanis, K.P.: Survey of
  unmanned aerial vehicles (uavs) for traffic monitoring. Handbook of unmanned
  aerial vehicles pp. 2643--2666 (2015)

\bibitem{larman2004agile}
Larman, C.: Agile and iterative development: a manager's guide. Addison-Wesley
  Professional (2004)

\bibitem{lyu2018fast}
Lyu, H., Yin, Y.: Fast path planning for autonomous ships in restricted waters.
  Applied Sciences  \textbf{8}(12), ~2592 (2018)

\bibitem{Magalhaes2018}
Magalh{\~{a}}es, J., Damas, B., Lobo, V.: Reinforcement learning: The
  application to autonomous biomimetic underwater vehicles control. {IOP}
  Conference Series: Earth and Environmental Science  \textbf{172},  012019
  (jun 2018). \doi{10.1088/1755-1315/172/1/012019}

\bibitem{mnih2015human}
Mnih, V., Kavukcuoglu, K., Silver, D., Rusu, A.A., Veness, J., Bellemare, M.G.,
  Graves, A., Riedmiller, M., Fidjeland, A.K., Ostrovski, G., et~al.:
  Human-level control through deep reinforcement learning. Nature
  \textbf{518}(7540), ~529 (2015)

\bibitem{shen2019}
Shen, H., Hashimoto, H., Matsuda, A., Taniguchi, Y., Terada, D., Guo, C.:
  Automatic collision avoidance of multiple ships based on deep q-learning.
  Applied Ocean Research  \textbf{86},  268 -- 288 (2019).
  \doi{https://doi.org/10.1016/j.apor.2019.02.020}

\bibitem{tran2013}
Tran, N.H., Choi, H.S., Baek, S.H., Shin, H.Y.: Tracking control of an unmanned
  surface vehicle. In: Zelinka, I., Duy, V.H., Cha, J. (eds.) AETA 2013: Recent
  Advances in Electrical Engineering and Related Sciences. pp. 575--584.
  Springer Berlin Heidelberg, Berlin, Heidelberg (2014)

\bibitem{vinyals2019grandmaster}
Vinyals, O., Babuschkin, I., Czarnecki, W.M., Mathieu, M., Dudzik, A., Chung,
  J., Choi, D.H., Powell, R., Ewalds, T., Georgiev, P., et~al.: Grandmaster
  level in starcraft ii using multi-agent reinforcement learning. Nature
  \textbf{575}(7782),  350--354 (2019)

\bibitem{wang2018path}
Wang, C., Zhang, X., Li, R., Dong, P.: Path planning of maritime autonomous
  surface ships in unknown environment with reinforcement learning. In:
  International Conference on Cognitive Systems and Signal Processing. pp.
  127--137. Springer (2018)

\bibitem{wang2014}
{Wang}, H., {Zhou}, J., {Zheng}, G., {Liang}, Y.: Has: Hierarchical a-star
  algorithm for big map navigation in special areas. In: 2014 5th International
  Conference on Digital Home. pp. 222--225 (Nov 2014).
  \doi{10.1109/ICDH.2014.49}

\end{thebibliography}

\end{document}